\documentclass[9pt,twocolumn]{extarticle}
\usepackage{hyperref}
\usepackage{amsfonts}
\usepackage{mathtools}
\usepackage{multicol}
\usepackage{xcolor}
\usepackage{multirow}
\usepackage{colortbl}
\usepackage{float}
\usepackage{booktabs}
\usepackage[labelfont=bf]{caption}
\usepackage{tabularx}
\usepackage{graphics}
\usepackage{adjustbox}
\usepackage{fancyhdr}
\usepackage{balance}

\title{\Huge \textbf{ Knowing When to Quit: Selective Cascaded Regression with Patch Attention for Real-Time Face Alignment}}

\usepackage{authblk}
\author[1,2]{Gil Shapira}
\author[1]{Noga Levy}
\author[1]{Ishay Goldin}
\author[1]{Roy J. Jevnisek}
\affil[1]{Samsung Semiconductor Israel R\&D Center (SIRC)}
\affil[2]{Faculty of Engineering, Bar-Ilan University}
\affil[ ]{\textit {ggiillsshhaappiirraa@gmail.com}}
\affil[ ]{\textit {nogaor@gmail.com}}
\affil[ ]{\textit {\{ishay.goldin,roy.jewnisek\}@samsung.com}}
\date{}

\begin{document}
\maketitle

\begin{abstract}
Facial landmarks (FLM) estimation is a critical component in many face-related applications.
In this work, we aim to optimize for both accuracy and speed and explore the trade-off between them. 
Our key observation is that not all faces are created equal. Frontal faces with neutral expressions converge faster than faces with extreme poses or expressions. To differentiate among samples, we train our model to predict the regression error after each iteration. 
If the current iteration is accurate enough, we stop iterating, saving redundant iterations while keeping the accuracy in check. We also observe that as neighboring patches overlap, we can infer all facial landmarks (FLMs) with only a small number of patches without a major accuracy sacrifice.
Architecturally, we offer a multi-scale, patch-based, lightweight feature extractor with a fine-grained local patch attention module, which computes a patch weighting according to the information in the patch itself and enhances the expressive power of the patch features. We analyze the patch attention data to infer where the model is attending when regressing facial landmarks and compare it to face attention in humans.
Our model runs in real-time on a mobile device GPU, with 95 Mega Multiply-Add (MMA) operations, outperforming all state-of-the-art methods under 1000 MMA, with a normalized mean error of 8.16 on the 300W challenging dataset. The code is available at \emph{\url{https://github.com/ligaripash/MuSiCa}}
\end{abstract}



\begin{figure}[H]
\includegraphics[width=0.96\linewidth]{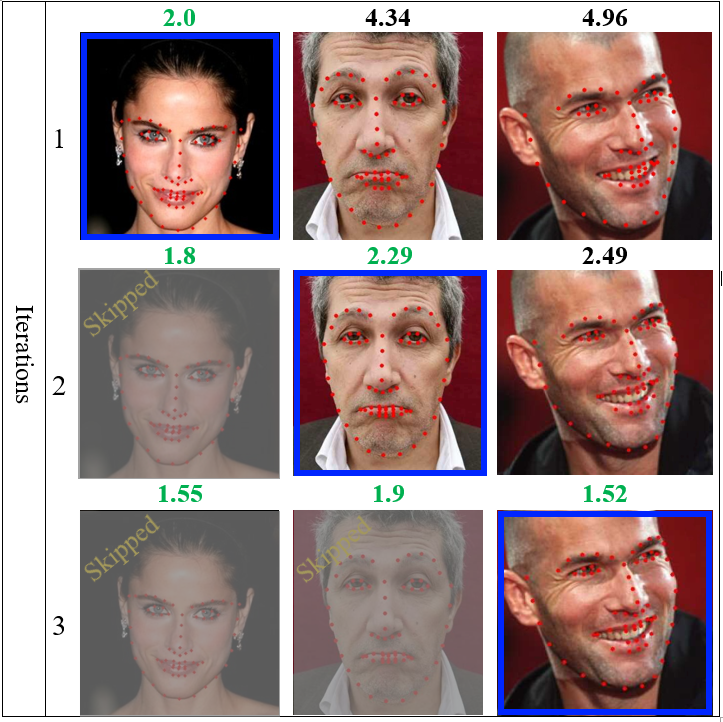}
\caption{Selectively choosing the number of iterations according to the estimated error. Each column presents the predictions of one image after iterations 1, 2 and 3. The estimated error is printed above each sample. In this example, the success threshold is set to 2.4. Estimated errors below this value are marked in green and the following iterations are skipped (grayed out), saving redundant computation. The final output of each sample is framed in blue.}
\label{fig:selective_iterations}
\end{figure}

\section{Introduction}
Face alignment, or facial landmarks regression, is a vital step in many face related applications, such as face recognition, detection of facial expressions, face beautification, avatar rendering, and more. Many of these applications need to run on mobile devices in real-time, but optimizing for speed has been largely overlooked in previous years. Most of the published facial alignment papers \cite{zheng2020hafanet,wu2018look,chen2019face, wang2020deep}, focus on accuracy optimization at the expense of enormous computation demands. The most accurate models achieve their accuracy at a price tag of more than 10 Giga Mutiply-Add operations (GMA), which is a far cry from real-time implementations on mobile devices. On the other end of the spectrum, regression-trees algorithms \cite{kazemi2014one, lee2015face} are super fast but lack the accuracy of CNN based algorithms.

In this work, we aim to close this gap, putting our attention on both accuracy and speed, and exploring the trade-off between them. We base our solution on the cascaded regression paradigm as it lends itself to computation optimization. The input to the model is cropped patches around the initial estimate of the location of the landmarks, instead of the full image. Each iteration in the process improves upon the estimate of the previous. All previous models in this paradigm had a fixed number of iterations. Our observation is that the input face complexity should determine the number of iterations: simple frontal faces can be regressed accurately with only one iteration.
Further accuracy increase in the next iteration would be marginal and may not justify the additional compute. On the other hand, faces with extreme poses, expressions, lighting conditions or occlusions may need all iterations to gain the required accuracy. This idea is illustrated in Figure \ref{fig:selective_iterations}. To implement it, we train our model to predict the regression error in each iteration. We then need to determine a threshold for skipping the next iterations. Every threshold value determines a specific operating point in the compute vs. accuracy domain. In this paper we explore this domain and show how to select a working point judiciously.

Architecturally, we take inspiration from MDM \cite{trigeorgis2016mnemonic} for its lightweight feature extractor with only two convolutions and effective information sharing between iterations. We improve their architecture in several ways. Firstly, we use a coarse-to-fine approach, cropping patches from a coarse face image on the first iteration, and progressively higher resolutions on later iterations. As patches size is fixed, each patch covers a large portion of the face in the first iteration, which assists the model to infer the correct location and pose of the face, and smaller portion in later iterations allowing the algorithm for better local precision. Secondly, prior cascaded regression models all had one to one relationship between facial landmarks (FLMs) and patches. Observing that neighboring patches overlap, we propose models regressing 68 FLMs with 34 or 19 predefined patches. These models reduce the computation load considerably, with only a small accuracy decrease. These 34 or 19 patches have proved to be sufficient also in localizing many more facial keypoints (100-300 in some applications) without any added computational cost. Thirdly, we improved the expressive power of the model by utilizing a fine-grained weighting for each component in the patch feature vector according to the patch feature itself, emphasizing salient information and suppressing irrelevant patch data.

Our main contributions include:
\begin{itemize}

    \item We expand the current cascaded regression paradigm of a fixed number of iterations by learning the expected regression error and thresholding it to get the appropriate number of iterations per sample.
    
    \item We relax the one to one relation between patches and landmarks in cascaded regression to reduce the computation load, with only a minor accuracy impact.
    
    \item We devise an iterative coarse-to-fine architecture with local patch attention to increase the model’s expressiveness while keeping computation demands in check.
    
    \item We offer a new data augmentation approach to reduce overfitting to the ‘mean face’ by identifying faces that are ‘far’ from the mean face and oversampling them. 
\end{itemize}

Utilizing these ideas, our multi-scale Selective Cascaded Regression with patch attention (MuSiCa) model achieves the best accuracy for models under 1 [GMA] operations on the 300W and WFLW datasets. We implement our model on mobile-phone GPU and it runs in real-time.


\section{Related Work}

There are two dominating approaches to face alignment in modern literature: direct regression and heatmap regression.

\textbf{Direct coordinates regression methods} regress the coordinates of the landmarks directly from the image or image patches. Most of these methods use the cascaded regression paradigm \cite{wu2017facial, mahpod2021facial, lee2015face, trigeorgis2016mnemonic, kazemi2014one, xiong2013supervised} which provides a good tradeoff between accuracy and speed. The algorithm starts from an initial FLMs configuration, usually the 'mean face'. On each iteration of the cascade, patches are cropped around the FLMs current estimated location. The model tries to infer the displacement from the initial position to the correct FLMs location, according to features extracted from the image patches. Each regressor in the cascade is trained to infer the correction step from the locations predicted by the previous regressor. ERT \cite{kazemi2014one} implements this concept by using an ensemble of regression trees with gradient boosting. SDM \cite{xiong2013supervised} extract SIFT features from local patches to estimate the FLMs coordinates. MDM \cite{trigeorgis2016mnemonic} extends SDM by replacing SIFT with lightweight CNNs and sharing information between iterations in the cascade. 

\textbf{Heatmap regression methods:} instead of regressing the landmarks coordinates, these methods regress one heatmap per landmark. The heatmap designates the probability of the landmark location in each pixel. These heatmaps are created by deep-stacked hourglass \cite{chen2019face, liu2019semantic, wu2018look} or U-net architectures \cite{dapogny2019decafa}, and achieve very high accuracy at the expanse of enormous computation resources. Most of them don't achieve real-time performance on strong GPUs, let alone mobile devices.

\textbf{Real-time deep methods} have been mostly overlooked in the past but gained some traction recently. \cite{liu2019efficient} purpose a two-stage network. The first network is lightweight, and only normalizes the input face image to a canonical pose, and the second more massive network regresses the landmarks on the normalized face. Another two-stage architecture is presented by \cite{duan2019faster}. They optimize for speed and achieve 1100 fps at the expanse of degraded accuracy. 

\textbf{Learning to attend} to the salient information is one of the essential concepts in deep learning, with applications in natural language processing \cite{wang2016attention, vaswani2017attention}, speech recognition \cite{chorowski2015attention, bahdanau2016end} and computer vision \cite{you2016image, mnih2014recurrent, parmar2019stand}. In the face alignment domain, attention mechanism is used in single-stage models to attend to the location of the landmarks selectively. \cite{yue2018attentional} offers a single-stage regression model with a multi-scale attention map learned by direct supervision to help the model attend to the location of the landmarks. Similarly, \cite{dapogny2019decafa} offers a cascade of U-nets with the same kind of supervised multi-scale attention maps. We inspect the attention maps learned by our model to better understand the facial features useful for this task, and offer an insightful comparison to human face attention.

\textbf{Error Estimation} papers are scarce in facial alignment literature inspite of its importance to downstream applications. \cite{kim2017local} offer two types of landmarks confidence measures: local and global. The local measure is inferred from the local feature of each landmark, and the global is computed from a 3D rendered face model. The motivation for the confidence measure is to assist the face recognition task. \cite{kumar2020luvli} assess landmarks localization uncertainty per landmark and also detect their visibility. In our work, the average normalized error is directly predicted for the whole face, as we are interested in fail/pass value for the complete regression for early stopping purposes.

\section{MuSiCa: Multi-scale Selective Cascaded Regression with Patch Attention}

\subsection{Solution Details}

\begin{figure*}
\includegraphics[width=\textwidth]{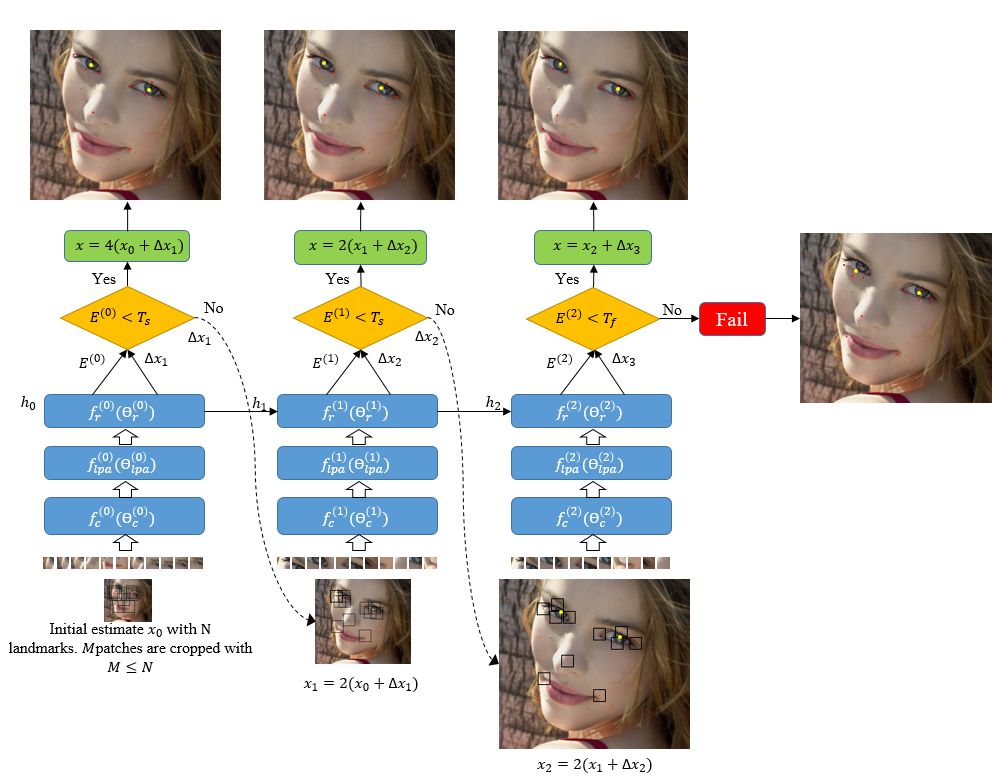}
\caption{An illustrative example of MuSiCa. We start with a coarse version of the input face and crop M small patches (14x14) at the location of the ‘mean face’. To save computation, the number of patches can be smaller than the number of landmarks. Landmarks without corresponding patches are marked in yellow. A lightweight CNN, $f_c^{(i)}(\cdot,\theta_c^{(i)})$ extracts features from the patches. The local patch attention module $f_{lpa}^{(i)}(\cdot, \theta_{lpa}^{(i)}) $ computes attention weights for each patch feature according to the patch feature itself. We aggregate the patch features to a single descriptor and pass it to the recurrent module $ f_r^{(i)}(\cdot, \theta_r^{(i)} ) $ that computes a hidden representation $\mathbf{h}(i)$ from the patch features of the current iteration and the hidden state of the previous iteration. The displacement $\mathbf{\Delta x_i}$ is computed from the hidden vector of each iteration. At the next iteration, we crop patches from a finer resolution according to the previously computed displacements. As patches are cropped from different scales, the weights are not shared between iterations. In addition to the estimated displacement, we compute the estimated error $E^{(i)}$ after each iteration. We use the estimated error to stop the computation if the accuracy is good enough $(E^{(i)} < T_s)$. If the estimated error after the last iteration is above the failure threshold $T_f$ then we declare a regression failure.}
\label{fig:arch-coarse}
\end{figure*}

\begin{figure*}
\includegraphics[width=\textwidth]{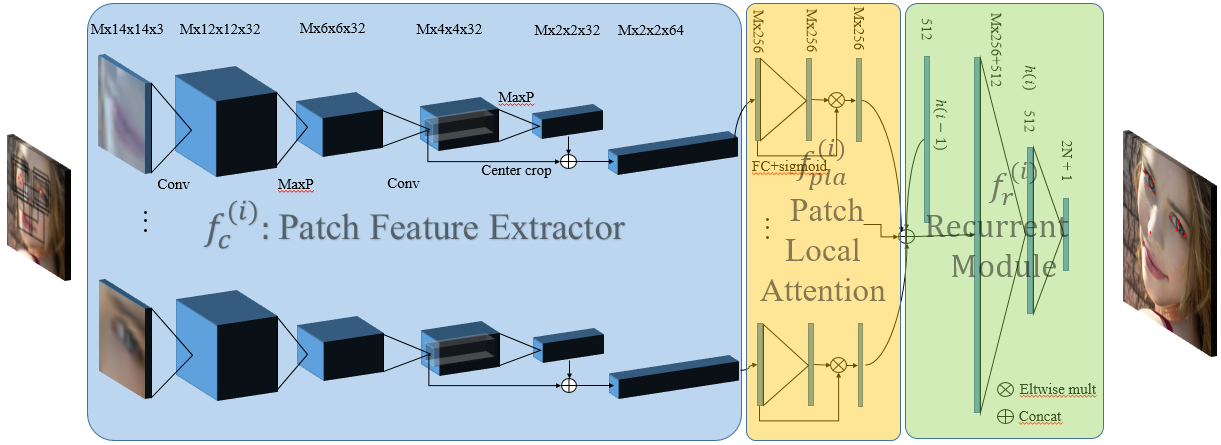}
\caption{An illustrative example of a single MuSiCa iteration. Small (14x14) patches are cropped around the current estimated landmarks location. We extract features from each patch using a lightweight CNN with two convolutions. Each patch feature is used to compute a fine grained, local patch attention vector using a fully-connected layer with sigmoid activation. The weight vector is later element-wise multiplied by the patch vector. The weighted features are all concatenated together and with the hidden state $h^{(i-1)}$ from previous iteration. The concatenated vector is used to compute the new hidden state $h^{(i)}$. The estimated $N$ landmarks and the error estimation $E^{(i)}$ are computed from the hidden state using a fully-connected layer.
}
\label{fig:single_iteration_arch}
\end{figure*}

Given a cropped face image $\bold{I} \in \mathbb R^{w \times h}$, face alignment is the task of localizing $N$ predefined landmarks $\mathbf{S} \in \mathbb R^{N \times 2}$. To save cycles, we take the multi-scale cascaded regression approach. In iteration $i$, the model receives the current estimate for the locations of the landmarks, $\mathbf{S}^{(i)}$, and a face image $I^{(i)}$. The image resolution doubles with every iteration. In the first iteration, $\mathbf{S}^{(0)}$ is set to an initial guess (the 'mean face') and $I^{(0)}$ to the lowest resolution image. In each iteration the model infers the displacement $\Delta \mathbf{x}^{(i)}$ from $\mathbf{S}^{(i)}$ to the ground truth as expressed in Eq. \ref{eq:1}

\begin{equation}
\mathbf{S}^{(i+1)} = 2\mathbf{S}^{(i)} + \Delta \mathbf{x}^{(i+1)}
\label{eq:1}
\end{equation}
The multiplication by 2 is due to the image up-scaling in the next iteration.
Our solution is illustrated in Figure \ref{fig:arch-coarse}.
Following MDM \cite{trigeorgis2016mnemonic}, our model contains a recurrent component that transfers information between iterations, assisting fast convergence. Unlike MDM, each iteration regresses images in different scales; hence we use separate parameterization for each iteration.
In each iteration, we feed our network with small (14x14) patches cropped around the landmarks. As our patches are fixed in size, they cover a large portion of the small face at the first iteration and smaller parts as the resolution increases, and localization improves, zooming in on the targets. Observing that some landmarks are densely packed (for instance, landmarks around the eyes) and patches overlap, we relax the standard paradigm in cascaded regression of cropping patches around each landmark. To reduce computation, we crop patches around $\mathbf{P} \subseteq \mathbf{S}$ landmarks skipping tightly packed patches. The patch-less landmarks are regressed using the information in patches of neighboring landmarks. Determining the size of $\mathbf{P}$ is a useful handle moving the model on the accuracy-computation tradeoff curve.
In Figure \ref{fig:arch-coarse}, we mark landmarks in $\mathbf{S} \setminus \mathbf{P}$ in yellow (only the pupils in this example) and landmarks in $\mathbf{P}$ in red. The cropped patches are aggregated and passed to a lightweight patch feature extractor $f_c^{(i)}$, where $i$ is the iteration index, see figure \ref{fig:single_iteration_arch}. The feature extractor is composed of two regular convolutions and max-pooling in between. To get the final multi-scale patch descriptor, we crop the center of the second convolution, and concatenate it with the second max-pool output.

\textbf{Local patch attention:} At the next stage, we compute a fine-grained patch attention function $f_{lpa}^{(i)}$ , inferring a single weight to each component of the patch feature vector, with a fully connected layer (FC) appended with a sigmoid activation to get weights between 0 and 1. The weight vector is later element-wise multiplied with the patch feature vector to produce the weighted feature vector.

\textbf{Recurrence}: To share information between iterations, We compute a hidden state vector $\textbf{h}^{(i)}$ using the function $f_r^{(i)}$. The input to $f_r^{(i)}$ is the concatenation of all patch vectors with the hidden state of the previous iteration $\textbf{h}^{(i-1)}$. $f_r^{(i)}$ is implemented with a FC layer followed by a tanh activation (Eq \ref{eq:2}, \ref{eq:3}).
\begin{equation}
\label{eq:2}
f^{(i)} :=  f_r^{(i)} \circ f_{lpa}^{(i)} \circ f_c^{(i)}
\end{equation}
\begin{equation}
\label{eq:3}
\mathbf{h}^{(i)} = f^{(i)}(\cdot, \mathbf{S}^{(i-1)}, \mathbf{\theta}^{(i)}, \mathbf{h}^{(i-1)})
\end{equation}
In Eq. \ref{eq:3}, $\theta^{(i)}$ denotes the weights of iteration network $f^{(i)}$.
To infer the final landmarks displacements, we use the function $f_l^{(i)}$. $f_l^{(i)}$ is implemented as another FC layer taking $\mathbf{h}^{(i)}$ as input (Eq. \ref{eq:4}). 
\begin{equation}
\label{eq:4}
\mathbf{\Delta x}^{(i)} = f_l^{(i)}(\mathbf{h}^{(i)}; \mathbf{\theta}_l^{(i)})
\end{equation}

\subsection{Error Estimation (Knowing When to Quit)}
In addition to the landmarks displacement, we compute the normalized estimated error $E(i)$ for each iteration (Eq. \ref{eq:5}), 
\begin{gather}
\label{eq:5}
E^{(i)} = f_e^{(i)}(\mathbf{h}^{(i)};  \mathbf{\theta}_e^{(i)})
\end{gather}
and compare the estimated error against a regression success threshold $T_s$. If $E(i) < T_s$, then the computation is terminated with the current iteration; Otherwise, it continues.  If the estimated error in the final iteration is above the failure threshold $T_f$ then we declare the output is invalid. Employing this early stopping scheme allows us to assign computation resources per sample as needed. Simple frontal faces can usually be regressed in one iteration while faces with extreme pose or expression may need all iterations (Figure \ref{fig:selective_iterations}).

\textbf{Choosing $\mathbf{T_s}$} allows us to control the accuracy-computation tradeoff without retraining the model. To judiciously choose a threshold, we would like to know the computation complexity and the expected error of the model induced by each $T_s$ value. For each sample, we select the number of iterations required to produce an estimated error below $Ts$; If this criterion is not met, we choose the last iteration.

\begin{figure}[H]
\includegraphics[width=\linewidth]{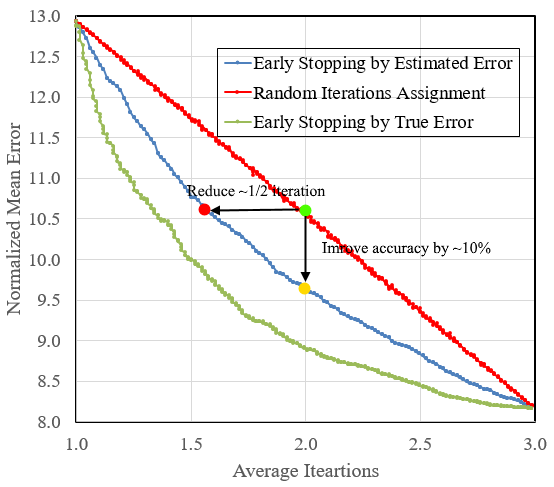}
\caption{The effectiveness of our early stopping policy. By selecting the number of iterations per sample according to our error estimator (blue graph), we can control the accuracy/time complexity tradeoff and significantly improve over random iteration selection (red graph). Selecting the number of iterations per sample according to the actual error (green graph) marks the best possible assignment}
\label{fig:iterations_vs_accuracy}
\end{figure}

After setting the required number of iterations per sample for a given $Ts$, we compute the average number of iterations and the normalized mean error for this policy (blue curve in Figure \ref{fig:iterations_vs_accuracy}). As the baseline to assess this policy's effectiveness, we count how many samples are assigned to each iteration and randomly assign iterations to samples with the same count. We repeat this random assignment 50 times to get a stable result (red curve in Figure \ref{fig:iterations_vs_accuracy}). These randomized policies have the same computation demand as the original policy, so the accuracy improvement is attributed to our policy's effectiveness. Relative to a point with an average of $2$ iterations on the baseline (green point), we improve the accuracy by $\sim10\%$ if we keep the average iterations constant (yellow point), or reduce the average number of iterations by approximately half while maintaining the same accuracy (red point). All samples are taken from the 300W challenging set.

The green line describes the theoretical optimal iteration selection policy, achieved by using the true posterior error, instead of the estimated error.

\subsection{Training and Loss function}

We train our model end-to-end.
Following \cite{feng2019rectified} we use the $L1$ Loss with rectification of small errors to mitigate the impact of ground truth labeling errors. To compute the estimated error we add a loss term which is the absolute value of the difference between the estimated error and the actual error.


\textbf{Data Augmentation} is crucial for training a robust facial alignment system that generalizes well. This is especially true when training sets contain only a few thousand samples, like the 300W training set, and the face bounding boxes are not consistent in size and location, which is a common situation. \cite{ feng2019mining} conducted a thorough study, evaluating how different augmentation strategies contribute to face alignment. One of their significant findings is that geometric transformations (rotation, shear) are more important than texture transformations. Building on their work, we apply the following augmentations: non-uniform scaling, shearing, flip, in-plane rotation, bounding box perturbation, color jetting, and gray-scale transformation as described in \cite{ feng2019mining}

\subsection{Data Balancing}

Many face alignment datasets suffer from unbalanced data where frontal faces with regular expressions dominate the sample population. This imbalance causes overfitting to standard pose and hurts the performance of the algorithm. To mitigate this issue,  \cite{feng2018wing} suggests a pose-based data balancing (PDB) strategy: They align each ground-truth shape in the training set to the mean shape by Procrustes Analysis with the mean shape as the reference shape. Next, they apply a PCA to the aligned shapes and project each shape to the first principal axis, associated with pose variation. Later, they create a histogram of the projected values. Small bins in this histogram are related to extreme poses that are under-represented in the dataset. They flatten this distribution by duplicating samples with rare poses. Our generalized data balancing method (GDB) extends this idea by trying to identify not only faces with extreme poses but also extreme expressions and other, more subtle idiosyncrasies. 

\begin{figure}[H]
\includegraphics[width=\linewidth]{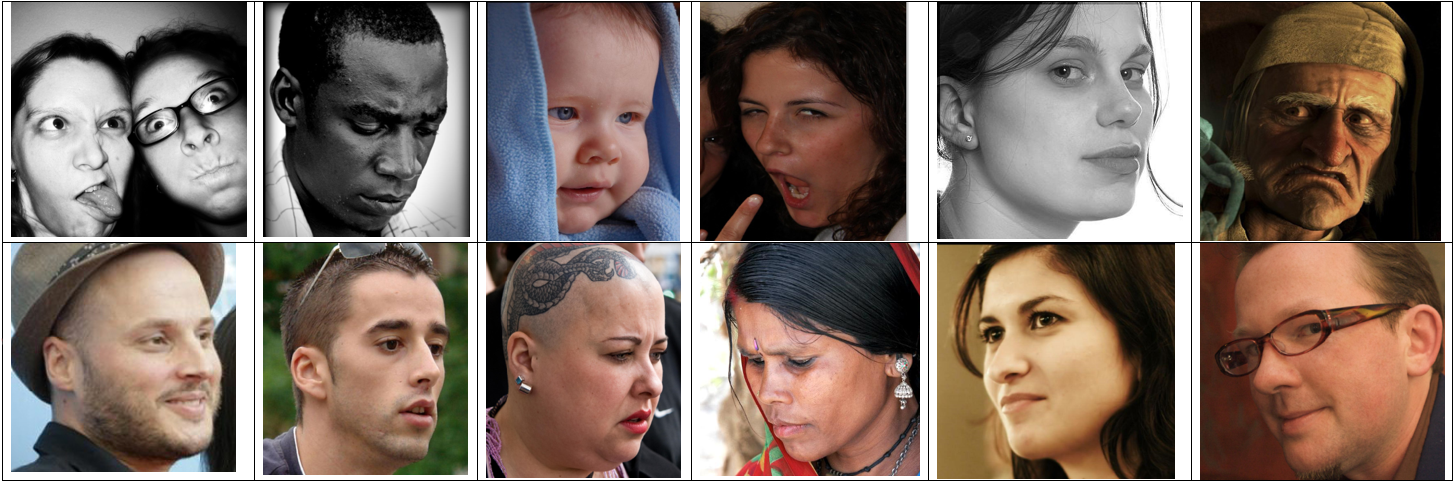}
\caption{Face variability induced by Mahalanobis distance vs. first principal axis. The top row depicts a random sample from the top 20 faces with the largest Mahalanobis distance out of 300W training set. They exhibit extreme values in pose, expression, age, etc. On the bottom row are six faces with extreme first principal axis values. This axis is related to pose only, and all other dimensions are ignored.}
\label{fig:mahal_vs_p0}
\end{figure}

We also perform a PCA to the aligned shapes, but instead of using only the first principal component, we take the first K components. Each shape is represented as a point in K-space. By measuring the Mahalanobis distance between a shape and the average shape, we quantify how well these shape features are represented in the training set. Shapes can have large Mahalanobis distance not only due to the pose, but also due to extreme expressions, age, or any other facial features that are under represented in the training set, as depicted in Figure \ref{fig:mahal_vs_p0}. The top row in this figure contains a random sample from 20 faces with the maximal Mahalanobis distance out of 300W training set. These samples have faces with extreme pose, extreme expressions, baby face etc. which are lacking in the dataset. The bottom row includes examples with extreme first principal axis values related to pose variability only. We over-sample faces in proportion to their Mahalanobis distance to enhance under represented samples.

\subsection{What are you looking at?}
\begin{figure}[H]
\includegraphics[width=\linewidth]{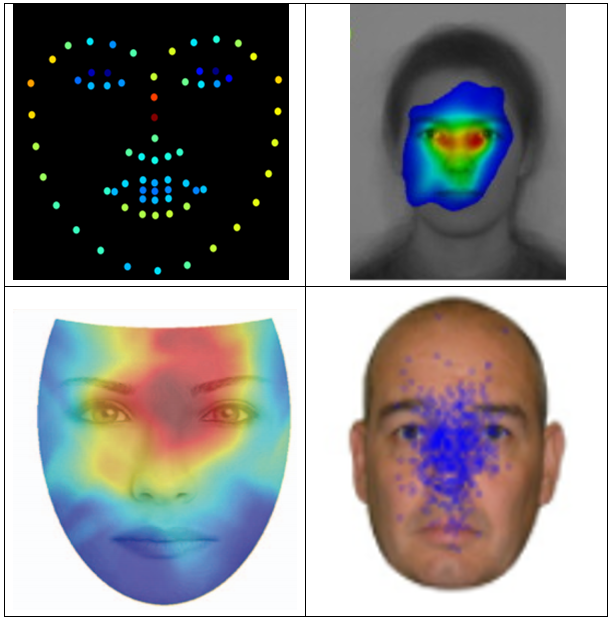}
\caption{The resemblance of patch attention in our model to human attention heatmaps. From upper-right clockwise: \cite{saether2009anchoring}, \cite{bindemann2009viewpoint}, \cite{zerouali2013optimal}, and our patch attention averaged over 300W common test set. The patch attention points are rendered at the FLMs initial position (the 'mean face'). In all cases the attention maxima is located on the upper part of the nose.}
\label{fig:path_attention}
\end{figure}

We were curious to know what part of the human face is the most indicative for the task of face alignment; in other words, which patches draw most of the model's attention. To answer this question, we extracted the patch attention vectors for all samples in the 300W-Fullset, averaged all 256 components of each vector, and then averaged over all samples in the set. The results are depicted in Figure \ref{fig:path_attention}. Most of the model's attention is concentrated around the upper center of the nose. This appeals to our intuition, as the nose is the most 'out-of-plane' organ in a face. As such, it exhibits the highest visual variability under out-of-plane rotations and is most indicative of a face pose.

We were also interested to know how the human visual system attends to the human face. \cite{zerouali2013optimal} answers this question by measuring the EEG signal, which appears 170 [ms] (N170) after the face stimulus, and is linked to early-stage face processing in the brain. \cite{bindemann2009viewpoint} and \cite{saether2009anchoring} support these findings by measuring the gaze time on each face location. 
Interestingly, As depicted in Figure \ref{fig:path_attention}, humans and our face alignment model, are all focusing their attention at the upper center of the nose when processing a face image. 

\begin{figure}[H]
\includegraphics[width=\linewidth]{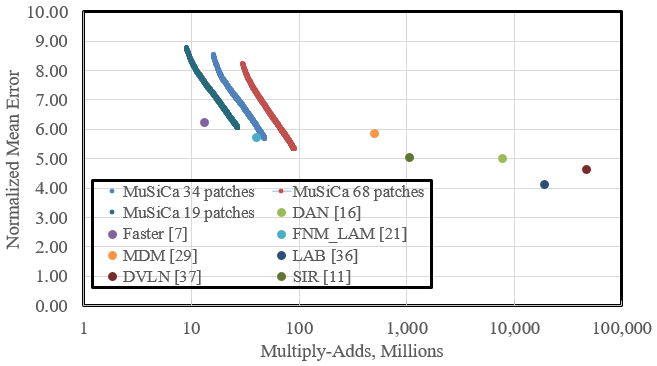}
\caption{Normalized mean error vs. Multiply-Adds count on 300W full test set. MuSiCa models create a segmented line in this space, that is induced by different early skipping thresholds. }
\label{fig:accuracy_vs_compute}
\end{figure}

\section{Experiments}
\subsection{300W}

300W \cite{sagonas2013300} is currently the most widely used dataset for face alignment. It is annotated semi-automatically with 68 landmarks, and also contains a bounding box for each face. The training set is composed of 3148 images, and the test set is composed of 689 faces, which are divided into the common set (554 faces mostly frontal or semi-frontal) and a challenging test set of 135 faces with extreme pose/expression/illumination. In our experiments, we put emphasis not only on the accuracy but also on the computation resources and the tradeoff between them. We normalize the errors by the inter-pupil distance as this is the common practice. The 300W-test results are summarized in Table \ref{table:300w}. The papers are divided into two groups: the green group with models less than 1[GMA] and the red group with models above 1[GMA]. Our MuSiCa68 model (MuSiCa with 68 patches) with 95 [MMA] dominate the green scoreboard, and outperforms much heavier models like MDM \cite{trigeorgis2016mnemonic} with 500[MMA], and DeFA \cite{liu2017dense} with 1.4 [GMA]. \\
Figure \ref{fig:accuracy_vs_compute} contains a graph of Normalized Mean Error vs. Multiply-Add operations for several models, measured on 300W full test set. Each one of the MuSiCa X models, where X is the number of patches, induces a segmented line in this space. Each point on this line is related to a different value of $T_{s}$ threshold value for early exit. This gives the freedom to choose the appropriate model according to the specific requirements of compute and accuracy without retraining the model. Table \ref{table:musica_complexity} contains the computation complexity of different MuSiCa models.

\begin{table}[]
\centering
\small
\begin{tabular}{@{}cccc@{}}

\toprule
\#FLMs & \#Patches & MA ops & Memory [MB]\\ \midrule
68     & 68        & 95     & 9.10\\
68     & 34        & 48     & 4.60\\
68     & 19        & 26     & 2.64\\
98     & 98        & 138    & 13.02\\
98     & 49        & 69     & 5.94\\ \bottomrule
\end{tabular}
\caption{MuSiCa variants Multiply Adds (Millions) with fixed 3 iterations}
\label{table:musica_complexity}
\end{table}

\begin{table}[H]
\scalebox{0.9}{
\begin{tabular}{@{}llllll@{}}
\toprule
\multicolumn{2}{l}{}                         & \multicolumn{4}{c}{300-W}                                                                                                                                            \\ \cmidrule(l){3-6} 
\multicolumn{2}{l}{\multirow{-2}{*}{Method}} & Common                               & Challenging                          & Fullset                              & MA                                              \\ \midrule
\multicolumn{2}{l}{LAB \cite{wu2018look}}                      & 3.42                                 & {\textbf{6.98}}                                 & 4.12                                 & \cellcolor[HTML]{FE0000}                        \\
\multicolumn{2}{l}{DCFE \cite{valle2018deeply}}                     & 3.83                                 & 7.54                                 & 4.55                                 & \cellcolor[HTML]{FE0000}                        \\
\multicolumn{2}{l}{DAN \cite{kowalski2017deep}}                      & 4.42                                 & 7.57                                 & 5.03                                 & \cellcolor[HTML]{FE0000}                        \\
\multicolumn{2}{l}{BL+AFS \cite{wang2020attention}}                   & {\textbf{3.13}} & 7.34 & {\textbf{3.82}} & \cellcolor[HTML]{FE0000}                        \\
\multicolumn{2}{l}{DVLN \cite{wu2017leveraging}}                     & 3.94                                 & 7.62                                 & 4.66                                 & \cellcolor[HTML]{FE0000}                        \\
\multicolumn{2}{l}{SIR \cite{hu2018facial}}                      & 4.29                                 & 8.14                                 & 5.04                                 & \cellcolor[HTML]{FE0000}                        \\
\multicolumn{2}{l}{TR-DRN \cite{lv2017deep}}                   & 4.36                                 & 7.56                                 & 4.99                                 & \cellcolor[HTML]{FE0000}                        \\
\multicolumn{2}{l}{DeFA \cite{liu2017dense}}                     & 5.37                                 & 9.38                                 & 6.10                                 & \cellcolor[HTML]{FE0000}                        \\ \midrule
\multicolumn{2}{l}{SDM \cite{xiong2013supervised}}                      & 5.60                                 & 17.00                                & 7.58                                 & \cellcolor[HTML]{0000FF}{\color[HTML]{000000} } \\
\multicolumn{2}{l}{CFAN \cite{zhang2014coarse}}                     & 5.50                                 & 16.78                                & 7.69                                 & \cellcolor[HTML]{0000FF}                        \\
\multicolumn{2}{l}{MDM \cite{trigeorgis2016mnemonic}}                      & 4.83                                 & 10.14                                & 5.88                                 & \cellcolor[HTML]{0000FF}                        \\
\multicolumn{2}{l}{FASTER \cite{duan2019faster}}                   & -                                    & -                                    & 6.25                                 & \cellcolor[HTML]{0000FF}                        \\
\multicolumn{2}{l}{FNM\_LAM \cite{liu2019efficient}}                 & 5.09                                 & 8.32                                 & 5.72                                 & \cellcolor[HTML]{0000FF}                        \\
\multicolumn{2}{l}{ERT \cite{kazemi2014one}}                      & -                                    & -                                    & 6.40                                 & \cellcolor[HTML]{0000FF}                        \\
\multicolumn{2}{l}{3DDFA \cite{zhu2016face}}                    & 6.15                                 & 10.59                                & 7.01                                 & \cellcolor[HTML]{0000FF}                        \\
\multicolumn{2}{l}{TCDCN \cite{zhang2014facial}}                    & 4.80                                 & 8.60                                 & 5.54                                 & \cellcolor[HTML]{0000FF}                        \\
\multicolumn{2}{l}{CFSS \cite{zhu2015face}}                     & 4.73                                 & 9.98                                 & 5.76                                 & \cellcolor[HTML]{0000FF} \\
\multicolumn{2}{l}{MuSiCa19 (ours)}                     & 5.22                                 & 9.50                                 & 6.05                                 & \cellcolor[HTML]{0000FF} \\
\multicolumn{2}{l}{MuSiCa34 (ours)}                     & 4.95                                 & 8.76                                 & 5.69                                 & \cellcolor[HTML]{0000FF} \\
\multicolumn{2}{l}{MuSiCa68 (ours)}               & {\textbf{4.63}} & {\textbf{8.16}} & {\textbf{5.32}} & \cellcolor[HTML]{0000FF}                        
\end{tabular}
}
\caption{Results on 300W testset. Errors are normalized by the inter-pupil distance. The blue models are below 1 [GMA] operations, and the red models are above this threshold. Our method with 68 patches is the most accurate in the light models category even though it has a complexity of only 90 [MMA], a far cry from the 1 [GMA] threshold.}\label{table:300w}

\end{table}

\subsection{WLFW}

\cite{wu2018look} created the WFLW dataset based on WIDER Face \cite{yang2016wider} dataset. Each sample contains 98 points annotations. There are 7500 training images and 2500 testing images which are separated to several subsets according to face attributes: large pose, expression, illumination, make-up, occlusion and blur. Table \ref{table:wflw} contains our results on this dataset. As before, we split the methods by computation complexity. Models above 1[GMA] operations are marked in red, and methods below this threshold are marked in green. Our MuSiCa98 model is the most accurate in the lightweight category. We follow the common practice for this dataset and normalized the errors by the inter-ocular distance.

\begin{table*}[]
\centering
\small
\begin{tabular}{@{}lcccccccl@{}}
\toprule
Method        & \multicolumn{1}{l}{Testset} & \begin{tabular}[c]{@{}c@{}}Pose\\ Subset\end{tabular} & \begin{tabular}[c]{@{}c@{}}Expression\\ Subset\end{tabular} & \begin{tabular}[c]{@{}c@{}}Illumination\\ Subset\end{tabular} & \begin{tabular}[c]{@{}c@{}}Make-Up\\ Subset\end{tabular} & \begin{tabular}[c]{@{}c@{}}Occlusion\\ Subset\end{tabular} & \begin{tabular}[c]{@{}c@{}}Blur\\ Subset\end{tabular} & MA                                              \\ \midrule
DVLN \cite{wu2017leveraging}  & 6.08                        & 11.54                                                 & 6.78                                                        & 5.73                                                          & 5.98                                                     & 7.33                                                       & 6.88                                                  & \cellcolor[HTML]{FE0000}                        \\
LAB \cite{wu2018look}  & 5.27                        & 10.24                                                 & 5.51                                                        & 5.23                                                          & 5.15                                                     & 6.79                                                       & 6.32                                                  & \cellcolor[HTML]{FE0000}                        \\
Wing \cite{feng2018wing}         & 5.11                        & 8.75                                                  & 5.36                                                        & 4.93                                                          & 5.41                                                     & 6.37                                                       & 5.81                                                  & \cellcolor[HTML]{FE0000}                        \\
HRNet  \cite{sun2019high}       & 4.60                        & 7.86                                                  & 4.78                                                        & 4.57                                                          & \textbf{4.26}                                            & 5.42                                                       & 5.36                                                  & \cellcolor[HTML]{FE0000}                        \\
AWing  \cite{wang2019adaptive}       & \textbf{4.36}               & \textbf{7.38}                                         & \textbf{4.58}                                               & \textbf{4.32}                                                 & 4.27                                                     & \textbf{5.19}                                              & \textbf{4.96}                                         & \cellcolor[HTML]{FE0000}{\color[HTML]{000000} } \\ \midrule
ESR \cite{cao2014face}          & 11.13                       & 25.88                                                 & 11.47                                                       & 10.49                                                         & 11.05                                                    & 13.75                                                      & 12.20                                                 & \cellcolor[HTML]{0000FF}                        \\
SDM \cite{xiong2013supervised}  & 10.29                       & 24.10                                                 & 11.45                                                       & 9.32                                                          & 9.38                                                     & 13.03                                                      & 11.28                                                 & \cellcolor[HTML]{0000FF}                        \\
CFSS \cite{cao2014face} & 9.07                        & 21.36                                                 & 10.09                                                       & 8.30                                                          & 8.74                                                     & 11.76                                                      & 9.96                                                  & \cellcolor[HTML]{0000FF}                        \\
MuSiCa44 (ours)     & 8.38                        & 17.89                                                 & 8.82                                                        & 8.12                                                          & 9.11                                                     & 10.6                                                       & 9.67                                                  & \cellcolor[HTML]{0000FF}                        \\
MuSiCa98 (ours)     & \textbf{7.90}               & \textbf{15.8}                                         & \textbf{8.52}                                               & \textbf{7.49}                                                 & \textbf{8.56}                                            & \textbf{10.04}                                             & \textbf{8.92}                                         & \cellcolor[HTML]{0000FF}                       
\end{tabular}
\caption{Results on WFLW. Models above 1[GMA] operations are marked in red, and models below this threshold are marked in blue. Both of our 98 and 44 patches models are more accurate than other models in the lightweight category.}
\label{table:wflw}
\end{table*}
\subsection{Implementation on a mobile phone}

We implemented a version of our model with 165 landmarks, 23 patches and, 2 iterations and without local patch attention on a Mali-G76 MP5 GPU of the Samsung A71 mobile phone. The measured run-time of our algorithm on this device is 14.77 [ms]. Samples of the output of this algorithm can be found in the supplementary material.

\section{Ablation Studies}

To assess the contribution of a specific aspect of our method, we create a model with the specific feature ablated and measure its accuracy on the 300W challenging test set. Ablating the local patch attention mechanism reduces the results by 8.72\% (see Table \ref{table:ablation}). Using PDB \cite{feng2018wing} for data balancing instead of our improved GDB reduced the accuracy by 3.3\%.
We also experimented with other architectures for our feature extractor. Specifically, we replace the regular convolutions layer with inverted bottlenecks \cite{sandler2018mobilenetv2}, but this model degraded the accuracy by 12.74\%. We also experimented with a global patch attention mechanism to complement our local patch attention. Global patch attention computes a weighting to each patch feature vector according to the data in other patches. We implemented a global attention module with both FC layer and a GCN \cite{kipf2016semi}, but both of them were below the baseline (degraded accuracy of 10.29\% and 12.50\%, respectfully)

\begin{table}[]
\small
\begin{tabular}{@{}lll@{}}
\toprule
Method                                                                                                             & \begin{tabular}[c]{@{}l@{}}300W \\ Challenging\\ Set\end{tabular} & \begin{tabular}[c]{@{}l@{}}Change \\ relative to \\ baseline (\%)\end{tabular} \\ \midrule
\begin{tabular}[c]{@{}l@{}}No local \\ patch attention\end{tabular}                                                & 8.72                                                              & 6.86                                                                           \\ \midrule
\begin{tabular}[c]{@{}l@{}}Trained with PDB \cite{feng2018wing}\\ data balancing \\ instead of \\ our GDB method\end{tabular} & 8.43                                                              & 3.31                                                                           \\ \midrule
\begin{tabular}[c]{@{}l@{}}Inverted bottlenecks\\  feature extractor\end{tabular}                                  & 9.20                                                              & 12.74                                                                          \\ \midrule
\begin{tabular}[c]{@{}l@{}}Global attention\\  with FC\end{tabular}                                                & 9.17                                                              & 10.29                                                                          \\ \midrule
\begin{tabular}[c]{@{}l@{}}Global attention\\  with GCN\end{tabular}                                               & {\color[HTML]{000000} 8.60}                                       & 12.37                                                                          \\ \midrule
MuSiCa68 (baseline)                                                                                                   & \textbf{8.16}                                                     & 0                                                                              \\ \bottomrule
\end{tabular}
\caption{Ablation experiments on the 300W challenging set. Both local patch attention and GDB contribute to increased accuracy. Our experiments with inverted bottlenecks \cite{sandler2018mobilenetv2}, and global patch attention did not improve the accuracy.}
\label{table:ablation}
\end{table}

\section{Conclusion}
In this paper, we study the tradeoff between accuracy and computation in face alignment by offering a cascaded regression method that selectively chooses when to quit the computation according to the estimated error, and by varying the number of patches used for landmarks regression. Using different thresholds for early skipping, we show how to choose a working point that meets specific accuracy and computation demands. To increase our model's expressiveness, we offer a local patch attention mechanism to highlight the important information and suppress the redundant patch data. We study what face part draws the model attention, compare it to human face attention, and draw parallels between them. We also offer an improved data balancing strategy and prove its effectiveness.
Our MuSiCa68 model offers an excellent tradeoff between accuracy and compute and achieves the best results for models under 1[GMA] on 300W and WLFW dataset. We implemented a version of our model on a mobile device GPU, and it runs in real-time.

\section{Acknowledgement}
We'd like to thank Geunhee Yang, Donghoon Kim, Yosi Keller, Tal Hassner, Ran Vitek and the paper reviewers for their constructive feedback and discussions.

\bibliographystyle{unsrt}
\balance
\bibliography{references}


\end{document}